RESEARCH ARTICLE

# Enhancing Automated Essay Scoring With Three Techniques: Two-Stage Fine-Tuning, Score Alignment, and Self-Training

**HONGSEOK CHOI[1], SERYNN KIM[2], WENCKE LIERMANN[1], JIN SEONG[1], AND JIN-XIA HUANG[1]**

[1]Electronics and Telecommunications Research Institute (ETRI), Daejeon 34129, South Korea
[2]Hankuk University of Foreign Studies, Seoul 02450, South Korea

Corresponding author: Hongseok Choi (hongking9@etri.re.kr)

This work was supported by the Institute of Information and Communications Technology Planning and Evaluation (IITP) grant funded by Korean Government through MSIT (Development of Semi-Supervised Learning Language Intelligence Technology and Korean Tutoring Service for Foreigners) under Grant RS-2019-II190004.

**ABSTRACT** Automated Essay Scoring (AES) plays a crucial role in education by providing scalable and efficient assessment tools. However, in real-world settings, the extreme scarcity of labeled data severely limits the development and practical adoption of robust AES systems. This study proposes a novel approach to enhance AES performance in both limited-data and full-data settings by introducing three key techniques. First, we introduce a Two-Stage fine-tuning strategy that leverages low-rank adaptations to better adapt an AES model to target prompt essays. Second, we introduce a Score Alignment technique to improve consistency between predicted and true score distributions. Third, we employ uncertainty-aware self-training using unlabeled data, effectively expanding the training set with pseudo-labeled samples while mitigating label noise propagation. We implement the above three key techniques on DualBERT. We conduct extensive experiments on the ASAP++ dataset, and additionally evaluate the proposed techniques on two other datasets, TOEFL11 and ELLIPSE, to examine their generalizability. In the *32*-data setting on ASAP++, all three key techniques improve performance, and their integration achieves 91.2% of the *full*-data performance trained on approximately 1,000 labeled samples. In addition, the proposed Score Alignment technique consistently improves performance in both *limited*-data and *full*-data settings: e.g., it achieves state-of-the-art results in the full-data setting on ASAP++ when integrated into DualBERT.

**INDEX TERMS** Automated essay scoring, deep learning, limited labeled data, low-resource setting, semi-supervised learning.

## I. INTRODUCTION

Automated Essay Scoring (AES) is a computer-based technology designed to automatically score essays [1]. With the recent growth of online learning platforms, large-scale assessments, and personalized learning systems, AES has been gaining increased attention [2].

AES systems can grade a large volume of essays rapidly; thus, it can significantly reduce workload of educators [3]. In addition, AES can provide consistent and immediate feedback to students, facilitating continuous learning without time constraints [4].

However, achieving reliable AES performance is challenging due to various factors such as the complex linguistic features of lengthy essays and different writing styles of novice learners. For a more detailed discussion of the challenges in AES, readers can refer to [5] and [6].

Over the past decade, task-specific methods utilizing deep learning, such as LSTM, CNN, and Transformer have been developed for AES [7], [8], [9]. More recently,

The associate editor coordinating the review of this manuscript and approving it for publication was Adamu Murtala Zungeru.

Large Language Models (LLMs) have achieved remarkable success across various NLP tasks [10], [11]. However, the performance of LLMs in AES has been relatively underwhelming. For example, the best performing LLM-based methods on the ASAP++ dataset achieve a Quadratic Weighted Kappa (QWK; an agreement measure between model-assigned and human-assigned scores, formally defined in Section IV-B) score of only around 0.5-0.6 [12], [13], [14], which is approximately 0.2 lower than that of task-specific deep learning models. Moreover, the substantial GPU requirements of LLMs pose a significant barrier to their adoption [15]. Consequently, task-specific deep learning models continue to be actively developed for AES [9], [16].

However, deep learning methods require a large amount of labeled data, which severely limits the practical application in real-world settings. One major reason is that data construction for AES presents several challenges [5]. First, labeling is labor-intensive and expensive. Second, maintaining consistency in human evaluations is difficult. Third, multi-trait AES, which evaluates aspects such as content and organization, often involves complex assessments based on different rubrics. Despite using concrete rubrics, assigned scores can vary significantly between raters. Here, *inter-rater agreement* refers to the degree of score consistency between independent human raters grading the same essays, and the 'overall' trait denotes a single holistic score that summarizes the quality of an essay, as opposed to analytic traits such as content and organization. Taghipour and Ng [17] reported a human inter-rater agreement in overall scoring with a QWK score of 0.754. Recently, many studies have explored multi-trait and cross-prompt AES, where the latter scores essays from unseen prompts [18], [19], [20], [21], [22]. However, few studies have addressed low-resource scenarios [23], [24], despite the significant challenges associated with constructing AES datasets.

In this study, we propose a novel approach to enhance AES performance in both limited- and full-data settings by introducing three key techniques: Two-Stage fine-tuning with low-rank adaptation (LoRA) [25], Score Alignment (SA), and Uncertainty-aware Self-Training (UST) [26]. LoRA is commonly used for parameter-efficient learning of LLMs, but we find that it can lead to further performance improvements even in smaller language models by capturing fine-grained features under our training strategy. We introduce a **Two-Stage fine-tuning** strategy in which LoRA layers are fine-tuned after training the base model. Next, to improve the consistency between predicted and true score distributions, we propose a **Score Alignment** technique that adjusts the predicted score distribution of the fine-tuned model to a more reliable distribution through a linear transformation. Furthermore, we employ **UST** for AES [26], which estimates the uncertainty of predictions on unlabeled data and uses this uncertainty in the training process. We filter out uncertain samples in the self-training phase to mitigate label noise propagation. In real-world educational scenarios, the main bottleneck in building AES datasets is the rating phase, whereas collecting unrated essays is relatively easy. To apply the above three key techniques, we build on DualBERT as our baseline model [16]. DualBERT captures both inter-token and inter-sentence semantics within an essay.

We conduct extensive experiments on the ASAP++ dataset, a widely used benchmark for AES [27]. Experimental results demonstrate the effectiveness of our approach. In the *full*-data setting (approximately 1,000 labeled samples for training), applying the Score Alignment technique to DualBERT led to a new state-of-the-art (SOTA) result, outperforming all previous AES methods. In the *32*-data setting, our three key techniques gradually improved performance, reaching 91.2% of the base model's *full*-data performance. Furthermore, additional experiments on two other datasets, TOEFL11 [28] and ELLIPSE [29], show similar trends—in particular, Score Alignment and UST improve performance on both datasets—demonstrating the generalizability of the proposed techniques. This study highlights the effectiveness of our AES approach in both few- and full-data settings.

To sum up, our contributions are as follows:

- We propose a novel AES approach that integrates three key techniques: Two-Stage fine-tuning, Score Alignment, and UST. The modularity of these techniques allows them to be easily combined with other methods, enabling incremental performance improvements particularly in low-resource settings.
- The proposed Score Alignment technique consistently improves performance in both limited-data and full-data settings. DualBERT combined with Score Alignment achieves new SOTA results, outperforming existing AES methods.
- We conduct extensive experiments and provide a detailed discussion on the three techniques: Two-Stage fine-tuning, Score Alignment, and UST.

The remainder of this paper is organized as follows. Section II reviews related work on AES. Section III formulates the task and describes the base model along with the three proposed techniques. Section IV details the datasets, evaluation metric, and experimental settings. Section V presents the results and discussion, Section VI discusses the limitations, and Section VII concludes the paper.

## II. RELATED WORK

### A. FEATURE-BASED

Early AES systems relied on manually engineered features such as essay length, grammar, syntax, and vocabulary usage, which were input into machine learning algorithms like regression models to predict essay scores [27], [30]. Recently, Li and Ng [21] demonstrated that a simple feature-based approach achieves SOTA results in overall scoring in cross-prompt settings.

### B. DEEP LEARNING-BASED

Over the last decade, various deep learning techniques, including LSTMs, CNNs, and Transformers, have been applied to AES [7], [17], [19], [31], [32], [33], [34], [35], [36], [37]. [18], [38] proposed a hybrid approach that combines deep learning techniques with linguistic features. Transformer-based pre-trained models, such as BERT and RoBERTa [39], [40], have demonstrated high performance in AES tasks [8], [9], [23], [41], [42]. Wang et al. [9] proposed a multi-scale essay representation method for BERT, encoding essays at token, segment, and document levels. Similarly, Cho et al. [16] introduced DualBERT, which encodes essays at both sentence and document levels. NPCR achieved SOTA performance in overall scoring using BERT [36]. NPCR predicts score differences between input essays and reference essays, and performs over 10 inference repetitions to ensure reliable performance. Yang et al. [8] proposed R2BERT, which incorporates a ranking loss term into the regression loss function in BERT.

### C. Seq2Seq-BASED

Recently, Seq2Seq models, such as T5 [43], have been employed for AES [20], [44]. Seq2Seq models decode hidden representations as token sequences. Do et al. [45] introduced AutoRegressive multi-Trait Scoring (ArTS), a T5-based model designed to generate multi-trait scores in the form of a text sequence. Chu et al. [46] generated rubric-guided rationales for each trait by using LLMs and appended these to the inputs of ArTS. Do et al. [44] enhanced the ArTS model by integrating Scoring-aware Multi-reward Reinforcement Learning (SaMRL).

### D. LLM-BASED

More recently, LLM-based AES approaches have been explored, mostly relying on simple prompting methods; their performance remains at QWK scores of only around 0.5–0.6 [12], [13], [14], [47], [48]. This is in contrast to the impressive results that LLMs have achieved in other NLP tasks [10], [11].

Many studies have focused on overall scoring and have often assumed the availability of sufficient data [9], [36], [44]. In this study, we address the multi-trait and both few- and full-data settings of AES and propose a novel deep learning-based approach.

Our three techniques can be positioned against prior work as follows. First, LoRA [25] was originally proposed for parameter-efficient fine-tuning of billion-parameter LLMs; we repurpose it for a BERT-scale multi-trait model as the second stage of a Two-Stage strategy that sweeps trait-specific loss-weight configurations (Section III-C). Second, whereas prior AES work improves score consistency *within* training—e.g., the ranking loss of R2BERT [8] or the reference-based prediction of NPCR [36]—our Score Alignment is a training-free post-processing step that corrects the distributional bias of already-predicted scores (Section III-D). Third, UST [26] was designed for few-shot text *classification*, where uncertainty is estimated from class probabilities; we adapt it to the AES *regression* setting by measuring uncertainty as the standard deviation of repeated-dropout predictions and by balancing the selected pseudo-labels over score bins instead of classes (Section III-E).

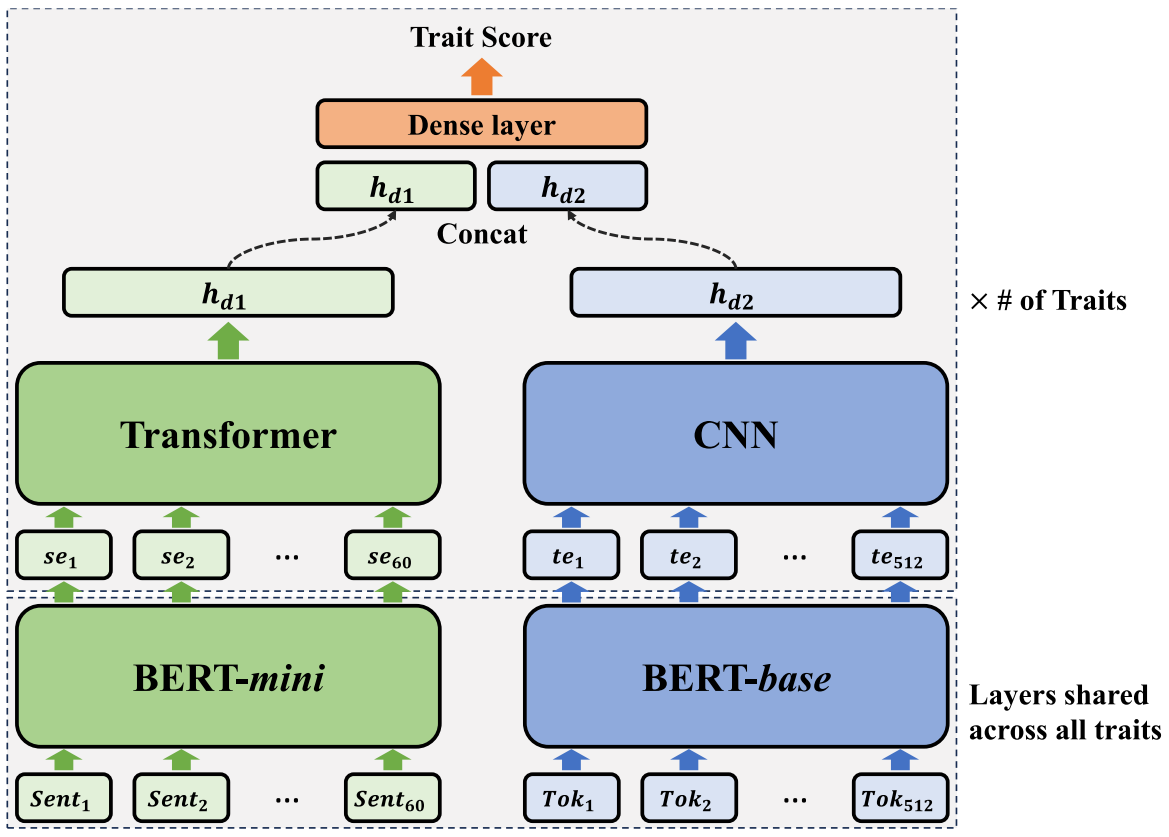

**FIGURE 1. Architecture of DualBERT for a single trait [16]. DualBERT has the same architecture across all traits except the 'overall' trait, where all $h_{d1}$ vectors from other traits are concatenated. The green and blue blocks indicate BERT-TransEnc and BERT-CNN, respectively.**

## III. PROPOSED METHOD

In this section, we first formulate the task and then describe the base model DualBERT and our three key techniques — Two-Stage fine-tuning, Score Alignment, and UST — which can be integrated into such base models.

### A. TASK FORMULATION

We address *multi-trait* essay scoring. Let $p$ denote a writing prompt and $\mathbb{T}_p = \{o, t_1, \ldots, t_m\}$ the set of traits defined for $p$, where $o$ denotes the 'overall' trait—the single holistic score summarizing the quality of an essay—and the remaining traits assess specific aspects of writing such as content and organization. A labeled dataset for $p$ is $D_p = \{(x_i, \mathbf{y}_i)\}_{i=1}^{n}$, where $x_i$ is an essay (a sequence of sentences) written for $p$, and $\mathbf{y}_i = (y_i^t)_{t\in\mathbb{T}_p}$ are its human-rated scores. Each $y_i^t$ is an integer on a predefined scale $[s_{\min}^t, s_{\max}^t]$ that depends on the dataset and, for ASAP++, on the prompt and trait (TABLE 1).

For training, every score is min-max normalized to $\tilde{y}^t = (y^t - s_{\min}^t)/(s_{\max}^t - s_{\min}^t) \in [0, 1]$. A scoring model takes the essay text $x$ as input and outputs the predicted normalized scores $\hat{\mathbf{y}} = (\hat{y}^t)_{t\in\mathbb{T}_p} \in [0, 1]^{|\mathbb{T}_p|}$ for all traits of the prompt. At inference time, each prediction is mapped back to the original scale by the inverse of this transformation and rounded to the nearest valid score, on which QWK is computed.

### B. DualBERT FOR MULTI-TRAIT SCORING

We build on DualBERT as our base model due to its high performance in multi-trait scoring [16]. This choice is supported by both evidence and design requirements: DualBERT achieves multi-trait performance on ASAP++ competitive with recent T5-based models while exhibiting lower variance (TABLE 3 and [16]), and, unlike Seq2Seq scorers that decode scores as text tokens, it outputs *continuous* scores through regression heads, on which our Score Alignment and uncertainty-aware self-training directly operate. An overview of the architecture is shown in Fig. 1. DualBERT comprises BERT-TransEnc for sentence- and document-level encoding and BERT-CNN for extracting local features at the document level. These representations are concatenated in the prediction layer and jointly optimized in a multi-task learning framework that considers multi-trait scores. BERT-TransEnc is designed to capture global features, such as the organization of writing, while BERT-CNN focuses on local features, such as grammatical errors.

More concretely, two pretrained encoders are shared across all traits: a BERT-mini encoder that encodes each sentence of the essay into a sentence vector, and a BERT-base encoder that encodes the token sequence of the whole essay. On top of these shared encoders, each trait $t \in \mathbb{T}_p$ has its own branch and prediction head. In BERT-TransEnc, a trait-specific Transformer encoder over the shared sentence vectors models inter-sentence relations and yields a document-level representation $h_{d1}$; in BERT-CNN, a trait-specific CNN over the shared BERT-base token representations extracts local features, yielding $h_{d2}$. The prediction head of each trait concatenates $[h_{d1}; h_{d2}]$ and applies a dense layer with a sigmoid activation to output the predicted normalized score $\hat{y}^t \in [0, 1]$. The 'overall' trait has its own branch of the same form, and its head input additionally concatenates the $h_{d1}$ vectors of all other traits, reflecting that the holistic score summarizes all aspects of an essay. The two shared encoders and the per-trait components account for the 120.7M and 13.8M$\times n_{\text{trait}}$ terms, respectively, of the parameter count reported in Section IV-D. The training loss is calculated as follows:

$$L = \alpha_o L_o + (1 - \alpha_o) \sum_{t_i \in \mathbb{T}_p \setminus \{o\}} \alpha_{t_i} L_{t_i}, \qquad (1)$$

where $\alpha_o$ and $\alpha_{t_i}$ denote the loss weights for the trait 'overall' and for the other trait $t_i$, respectively, and $\mathbb{T}_p$ denotes the trait set defined in Section III-A. Each loss term in Eq. (1) is the mean squared error (MSE) computed on the min-max normalized scores of Section III-A: for a trait $t \in \mathbb{T}_p$, $L_t = \frac{1}{|B|} \sum_{i \in B} (\hat{y}_i^t - \tilde{y}_i^t)^2$ over a mini-batch $B$, where $L_o$ is the MSE loss for the 'overall' trait and $L_{t_i}$ is that for the other trait $t_i$.

### C. TWO-STAGE FINE-TUNING WITH LoRA

To capture fine-grained features, we introduce a Two-Stage fine-tuning strategy using LoRA [25]. In the first stage, we fine-tune DualBERT, and then insert LoRA layers into the fine-tuned DualBERT model. In the second stage, we fine-tune the LoRA layers while the layers of the base model (i.e., DualBERT) are frozen. During this stage, we repeat the initialization-and-training process across traits and apply different loss weights for each trait every process. Our motivation is that, in the second stage, LoRA can find the best-performing model by sweeping various combinations of loss weights for traits, just as LoRA was originally designed to adapt LLMs to various target tasks. Although BERT-scale models can be fully fine-tuned on a modest GPU, we adopt LoRA not to save memory but to make this per-trait search modular: each configuration trains only the adapter layers on a single frozen base model, so sweeping many trait-specific loss-weight combinations is cheaper than fully fine-tuning a separate model for each. In the end, the parameters with the best performance on the development set are selected. The algorithm is described in Alg. 1. For easier understanding, we specify the settings for the loss weights $\alpha$ of Eq. (1) used in this study in lines 7–13; we apply the same settings to the three datasets (ASAP++, TOEFL11, and ELLIPSE). However, the weight settings are not restricted to a specific trait configuration—even without the 'overall' trait—and can be flexibly adjusted as hyperparameters.

**Algorithm 1** Two-Stage Fine-Tuning With LoRA

1: Initialize a base model $M_{\text{base}}$
2: Fine-tune $M_{\text{base}}$ on $D_{\text{train}}$ using $D_{\text{dev}}$ for validation
3: Freeze all the layers of $M_{\text{base}}$
4: $M_{\text{lora}} \leftarrow$ Insert LoRA layers to $M_{\text{base}}$
5: **for** target **in** {balance, overall, ..., voice} **do**
6: Initialize the LoRA layers
7: **if** target == 'balance' **then**
8: $\alpha_o = 0.7$, and $\alpha_{t_i} = 1.0 \, \forall$ traits
9: **else if** target == 'overall' **then**
10: $\alpha_o = 0.9$, and $\alpha_{t_i} = 0.1 \, \forall$ traits
11: **else**
12: $\alpha_o = 0.1$, $\alpha_{\text{target}} = 1.0$, and $\alpha_{t_i} = 0.1$ for the other traits
13: **end if**
14: Fine-tune $M_{\text{lora}}$ on $D_{\text{train}}$ using $D_{\text{dev}}$ for validation
15: $QWK_{\text{dev}} \leftarrow$ evaluate($M_{\text{lora}}, D_{\text{dev}}$)
16: **if** $QWK_{\text{dev}}$ is the best so far **then**
17: Save the parameters of $M_{\text{lora}}$
18: **end if**
19: **end for**

Formally, LoRA reparameterizes a pretrained weight matrix $W \in \mathbb{R}^{d \times k}$ of the frozen base model as $W' = W + \frac{\alpha_{\text{lora}}}{r} BA$, where $B \in \mathbb{R}^{d \times r}$ and $A \in \mathbb{R}^{r \times k}$ (typically with $r \ll \min(d, k)$) are the only trainable matrices, $\alpha_{\text{lora}}$ is a scaling factor, and $B$ is initialized to zero so that $BA = 0$ at the beginning of the second stage. LoRA modules are attached to the query, key, and value projection matrices of every attention layer in DualBERT—in the two shared BERT encoders and in the trait-specific inter-sentence Transformer encoders (Section III-B), which accounts for the 118M LoRA parameters reported in our setting—while all original parameters, including the CNNs and the prediction heads, remain frozen.

The loss-weight values in lines 7–13 were determined empirically in preliminary experiments. However, the final performance does not hinge on these specific values: since the second stage keeps only the configuration that performs best on the development set (lines 14–18), they act as a search space for the sweep rather than as fixed constants. The sensitivity of these and the other key hyperparameters is analyzed quantitatively in Section V-H.

In Alg. 1, 'balance' is not a scoring trait but a weighting configuration that keeps the balanced loss weights of the first stage. 'Overall' is the holistic trait; although it is one of the ASAP++ scores, it acts as a summary target correlated with every trait—its prediction head aggregates the representations of all other traits as described in Section III-A—so Eq. (1) assigns it a dedicated weight $\alpha_o$ that balances the holistic objective against the analytic-trait losses.

### D. SCORE ALIGNMENT

A fine-tuned model can inherit bias from the training data. For example, in a regression task with a score range of [0, 1], it tends to predict test samples with a true score of 1 as 0.95–0.99 rather than exactly 1; similarly, a true score of 0 as 0.01–0.05 rather than exactly 0. To address this issue, we propose a Score Alignment technique that adjusts the predicted score distribution on the test set using the prediction results on the development set. Since the fine-tuned model yields the best results on the development set, this adjustment can align the test set's predictions with a more reliable distribution by leveraging those on the development set.

Score Alignment is a post-processing technique that applies a linear transformation to the predicted scores of test samples. Rather than an ad-hoc or manually tuned adjustment, it is a principled, data-driven bias correction: the parameters $(a, b)$ of the transformation in Eq. (2) are estimated solely from the discrepancy between the true and predicted scores on the held-out development set. It is applied after training as a separate step—it introduces no trainable parameters, uses no test labels, and leaves the fine-tuned model weights unchanged. The formula is as follows:

$$\begin{aligned}\hat{y}_{\text{test, aligned}} &= \text{ScoreAlignment}(\hat{y}_{\text{test}}, y_{\text{dev}}, \hat{y}_{\text{dev}})\\ &= \text{LinearTransformation}(\hat{y}_{\text{test}}, a, b)\\ &= \frac{\hat{y}_{\text{test}} - \hat{y}_{\text{test, min}}}{\hat{y}_{\text{test, max}} - \hat{y}_{\text{test, min}}} \cdot (b-a) + a,\\ \text{where}\quad a &= \text{Clip}_0^1(\mu(y_{\text{dev, bottom-p\%}})\\ &\quad - \mu(\hat{y}_{\text{dev, bottom-p\%}}) + \hat{y}_{\text{test, min}}),\\ \text{and}\quad b &= \text{Clip}_0^1(\mu(y_{\text{dev, top-p\%}})\\ &\quad - \mu(\hat{y}_{\text{dev, top-p\%}}) + \hat{y}_{\text{test, max}}),\end{aligned} \tag{2}$$

where $y$ denotes the true labels (i.e., the true scores of essays), $\hat{y}$ denotes the predicted scores, $\text{Clip}_0^1$ is a clipping function that restricts values to the range [0, 1], and $\mu(y)$ denotes the mean function, defined as $\frac{1}{n_y}\sum_{i=1}^{n_y} y_i$. The subsets $y_{\text{top-p\%}}$ and $y_{\text{bottom-p\%}}$ represent the top and bottom $p\%$ of $y$, respectively, where $y_{\text{top-p\%}} = \{y_i \in y \mid y_i \geq q_{100-p}\}, y_{\text{bottom-p\%}} = \{y_i \in y \mid y_i \leq q_p\}$, and $q_p = \text{Quantile}(y, p\%)$; $p$ is a hyperparameter.

**TABLE 1. Summarization of the ASAP++ dataset. In Traits, Ovrl, Cnt, Org, WC, SF, Cnv, Nar, PA, Lng denote overall, content, organization, word choice, sentence fluency, conventions, narrativity, prompt adherence, language, respectively. Range denotes the original integer score scale of 'overall' / the other traits. (Data:** `https://www.kaggle.com/c/asap-aes/data`**.)**

| Prompt | #Essays | Avg.W | Type | Traits | Range |
|---|---|---|---|---|---|
| P1 | 1,783 | 350 | Persuas. | Ovrl/Cnt/Org/WC/SF/Cnv | 2–12/1–6 |
| P2 | 1,800 | 350 | Persuas. | Ovrl/Cnt/Org/WC/SF/Cnv | 1–6/1–6 |
| P3 | 1,726 | 150 | Src-based | Ovrl/Cnt/Nar/PA/Lng | 0–3/0–3 |
| P4 | 1,772 | 150 | Src-based | Ovrl/Cnt/Nar/PA/Lng | 0–3/0–3 |
| P5 | 1,805 | 150 | Src-based | Ovrl/Cnt/Nar/PA/Lng | 0–4/0–4 |
| P6 | 1,800 | 150 | Src-based | Ovrl/Cnt/Nar/PA/Lng | 0–4/0–4 |
| P7 | 1,569 | 250 | Narrat. | Ovrl/Cnt/Org/Cnv/Style | 0–30/0–6 |
| P8 | 723 | 650 | Narrat. | Ovrl/Cnt/Org/WC/SF/Cnv/Voice | 0–60/0–12 |

The core components of this formula are $a$ and $b$, which represent the minimum and maximum values of the transformed scores, respectively. The Score Alignment technique calculates the difference between the true and predicted scores within the range defined by the low and high scores on the development set. This difference is then added to the lower and upper bounds of the predicted scores on the test set, respectively. This mechanism aims to reduce distortions in predictions caused by bias in the fine-tuned model. Here, $a$ and $b$ are clipped to [0, 1] because the model outputs scores normalized to [0, 1], so any value outside this range is invalid.

### E. UNCERTAINTY-AWARE SELF-TRAINING

Uncertainty-aware Self-Training (UST), proposed by Mukherjee and Awadallah [26], was originally introduced for limited-data text classification. In contrast, we address AES, which is a regression task, and therefore apply UST with certain modifications. Specifically, we estimate the uncertainty of unrated essays by measuring it as the standard deviation of predictions obtained through repeated dropout operations. The predictions are obtained using a model trained on few labeled data. This repeated-dropout procedure corresponds to Monte Carlo dropout (MC-dropout). The uncertainty $\sigma_u$ can be formulated as shown in Eq. (3).

$$\sigma_u = \sqrt{\frac{1}{T}\sum_{i=1}^{T}\left(\hat{y}_{i,\text{dr}} - \frac{1}{T}\sum_{j=1}^{T}\hat{y}_{j,\text{dr}}\right)^2}, \tag{3}$$

where $T$ denotes the number of dropout repetitions, which is a hyperparameter, $\hat{y}_{i,\text{dr}}$ denotes the predicted scores obtained with dropout `dr` applied.

To maintain a balanced distribution of predicted scores, we partition the samples into $n_b$ bins based on the magnitude of their predicted scores. From each bin, we select $n_s$ samples with the lowest uncertainties. Thus, UST can effectively filter out noisy pseudo-labels with high uncertainties. These predicted scores are subsequently used as pseudo-labels, resulting in a total of $n_b \times n_s$ pseudo-labeled samples. Finally, we augment the original training set with the pseudo-labeled data and train a newly initialized model on the augmented dataset.

**TABLE 2. Statistics of the three datasets. In Category, P, S, and N denote persuasive, source-based, and narrative essays; ELL denotes English language learners. For ASAP++, the score scale is predefined and varies by prompt and trait.**

| | ASAP++ | TOEFL11 | ELLIPSE |
|---|---|---|---|
| **#Essays** | 13k | 12.1k | 6.5k |
| **#Prompts** | 8 | 8 | 44 |
| **#Traits** | 11 | 1 | 7 |
| **Category** | P,S,N | P | P |
| **Language** | Native | ELL | ELL |
| **Score scale** | varies | low/med/high | 1–5 |

### F. INTEGRATION OF THE THREE TECHNIQUES

A key advantage of the proposed three techniques is their modularity; that is, each can be applied independently. We combine these three because they address complementary bottlenecks of low-resource multi-trait AES: Two-Stage fine-tuning with LoRA adapts the model by searching over trait-specific loss-weight combinations, Score Alignment corrects the train-to-test distribution bias of the predicted scores, and UST leverages unrated essays via pseudo-labeling to enlarge the effective training set. Because they operate on different axes—model adaptation, output calibration, and data augmentation—they can be combined with little interference, and their individual and partial-combination contributions are complementary. This section outlines the integration process for the three techniques. First, after training the base model, DualBERT, we integrate LoRA and fine-tune the model. Next, Score Alignment is applied to obtain more reliable pseudo-scores on the unlabeled data. Then, UST is used to filter out noisy pseudo labels, and DualBERT is retrained on the resulting augmented dataset. Finally, Score Alignment is applied once more to refine the predictions. We denote this integration as 'LoRA+SA+UST'.

## IV. EXPERIMENTAL SETUP

### A. DATASETS

The Automated Student Assessment Prize++ (ASAP++) dataset is a widely used benchmark dataset in the field of AES, provided by a Kaggle competition [27]. TABLE 1 summarizes the ASAP++ dataset. It contains 12,978 essays written by students in grades 7–10, covering eight prompts across three types: persuasive, source-based, and narrative essays. Each essay was holistically scored by at least two human raters, with scores reflecting various writing traits such as content, organization, and sentence fluency. The essay lengths range approximately between 150 and 650 words. The ASAP++ dataset is useful for testing AES models in diverse scenarios such as multi-trait and cross-prompt settings.

To examine the generalizability of the proposed techniques, we additionally evaluate them on two datasets, TOEFL11 [28] and ELLIPSE [29]; TABLE 2 summarizes their statistics alongside ASAP++. TOEFL11 contains 12,100 independent-writing essays by non-native speakers from 11 first-language backgrounds, balanced across eight topics and annotated with a single holistic ('overall') score on a discrete three-level scale (low, medium, high). ELLIPSE consists of 6,482 essays by English language learners across 44 prompts, annotated with an 'overall' score and six analytic traits—cohesion, syntax, vocabulary, phraseology, grammar, and conventions—on a 1–5 scale. The two datasets differ from ASAP++ in writer population (native vs. ELL), score scale, and trait set.

### B. EVALUATION METRIC

The quadratic weighted kappa (QWK) is the official evaluation metric for AES. QWK measures agreement between human-rated scores (i.e., gold scores) and model-predicted scores. First, a weight matrix $W$ is calculated using:

$$W_{i,j} = \frac{(i-j)^2}{(N-1)^2} \tag{4}$$

where $i$ is the gold score, $j$ is the predicted score, and $N$ is the total score range. Then, a confusion matrix $O$ (observed) and an expected matrix $E$ (based on score distributions) are constructed. QWK is calculated as:

$$k = 1 - \frac{\sum_{i,j} W_{i,j} O_{i,j}}{\sum_{i,j} W_{i,j} E_{i,j}} \tag{5}$$

DualBERT outputs prediction scores in the range $[0, 1]$. At inference, we apply the inverse of the min-max normalization used during training to map each prediction back to the original score scale of the corresponding dataset, round it to the nearest valid score, and compute QWK on that scale. The original scale differs by dataset: for ASAP++ it is predefined and varies by prompt and trait; for ELLIPSE it is a 1–5 scale; and for TOEFL11 the prediction is first scaled to $[1, 5]$ and then mapped to three ordinal levels (low, medium, high) using thresholds of 2.25 and 3.75, following the official setting [28].

### C. EXPERIMENTAL SETTINGS

On ASAP++, we conduct extensive experiments in both full-data and $K$-data settings. For the full-data setting, we report the average results over five random seeds for train/dev/test splits with a 3:1:1 ratio. This yields approximately 1,000 training data points for each prompt (except Prompt 8). In the $K$-data setting, we sample $K$ essays for each of the training and development sets with a balanced score distribution, and the test set is the same as the one used in the full-data setting and the remaining data are used as unlabeled data. Because the test set is excluded from this unlabeled pool, our UST operates in an inductive rather than a transductive setting. This pool consists of the remaining unrated essays of the same prompt, which share the same rubric and, by construction, the same score distribution. As noted in Section I, the main bottleneck in real-world settings lies in rating essays rather than collecting them; UST is therefore well suited to scenarios such as grading a large volume of student essays. We distinguish two training settings with respect to prompts. In the single-prompt (SP) setting, a model is trained separately on each prompt, while in the multi-prompt (MP) setting, essays from all prompts are trained together. Unless otherwise noted, all experiments on ASAP++ were conducted in the SP setting.

In the MP setting, a single DualBERT is trained on the union of all prompts: the trait-specific branches and prediction heads cover the union of the trait sets of all prompts (eleven traits for ASAP++, one for TOEFL11, and seven for ELLIPSE), while the two BERT encoders are shared, exactly as in the SP setting (Section III-B). Because the prompts on ASAP++ have different trait sets (TABLE 1), traits that are not defined for a given essay's prompt are masked out, i.e., excluded from the loss in Eq. (1) for that essay. Mini-batches are drawn by uniformly shuffling the combined training set, and scores are min-max normalized per prompt and trait.

We also evaluate on TOEFL11 and ELLIPSE in both the full-data and $K$-data settings, using the same protocol as [15]. For these two datasets, we evaluate on each dataset's official test set, report the average over ten folds (each corresponding to a different train/dev split), and train in the MP setting by combining all prompts within each dataset.

### D. IMPLEMENTATION DETAILS

#### 1) DualBERT

The hyperparameter settings for our base model, DualBERT, are as follows. For training, AdamW was applied with a learning rate of 2e-5 and an epsilon of 1e-8 [49], the dropout rate was set to 0.1 [50], the batch size to 16, and the patience to 20 within a maximum of 100 training epochs. For the model architecture, BERT-mini was used for BERT-TransEnc, while BERT-base was used for BERT-CNN. Specifically, we used the publicly released `bert-base-uncased` checkpoint for BERT-base and the `prajjwal1/bert-mini` checkpoint (four layers, hidden size 256) for BERT-mini. Accordingly, "Initialize a base model" in line 1 of Alg. 1 means that the two shared BERT encoders are loaded from these pretrained checkpoints, whereas all remaining components of DualBERT—the trait-specific inter-sentence Transformer encoders, CNNs, and prediction heads (Section III-B)—are randomly initialized. The maximum number of input sentences for BERT-TransEnc was set to 60, and the input tokens for BERT-CNN were truncated to 512. A sigmoid function was used in the final dense layer, and the mean squared error (MSE) loss was applied. All the essay scores are min-max normalized for training. In Eq. (1), the loss weight $\alpha_o$ for 'overall' was set to 0.7, while the loss weights $\alpha_{t_i}$ for all other traits were set to 1.0.

#### 2) TWO-STAGE FINE-TUNING WITH LoRA

For LoRA, both `lora_r` and `lora_alpha` were set to 512 with a dropout rate of 0.05. LoRA layers were applied to the `query`, `key`, and `value` modules in BERT. In the second fine-tuning stage, the different loss weights were set depending on the 'target' as shown in Alg. 1. For 'balance', $\alpha_o = 0.7$ and $\alpha_{t_i} = 1.0$ for all other traits, which is the same as in the first fine-tuning stage. For 'target'='overall', $\alpha_o = 0.9$ and $\alpha_{t_i} = 0.1$ for all traits. If the 'target' is one of other traits, $\alpha_o = 0.1$, $\alpha_{target} = 1.0$, and $\alpha_{t_i} = 0.1$ for the remaining traits.

#### 3) SCORE ALIGNMENT AND UST

For Score Alignment, the $p\%$ in Eq. (2) was set to 5. The score alignment technique is applied for each trait. In the $K$-data settings, UST utilizes all essays, except for $2K$ samples and test data, as unlabeled data. The uncertainty is measured across all traits and then averaged. For the MC-dropout uncertainty estimate in Eq. (3), at inference we keep the standard dropout layers of the BERT encoders active (the attention-probability and hidden-state dropouts, with the same dropout rate of 0.1 as used during training) and take $T = 10$ stochastic forward passes; the per-sample standard deviation over these passes is used as the uncertainty. As described in Section III-E, $n_b$ and $n_s$ were set to 8 and 32, respectively. Therefore, the size of the augmented training set is $8 \times 32 + K$. We conduct the self-training process only one time without iterations.

#### 4) COMPUTATIONAL RESOURCES

We used NVIDIA A6000 GPU with 48 GB memory for training. In our setting, the DualBERT model has approximately $(120.7 + 13.8\,n_{\text{trait}})$M parameters, where $n_{\text{trait}}$ is the number of traits (e.g., about 273M for the eleven ASAP++ traits in the MP setting), while the LoRA layers have 118M parameters. When stored on a hard disk, DualBERT and LoRA occupy approximately 1GB and 0.45GB, respectively. For the implementation, we utilized Python 3.10 along with the Transformers (v4.39) and PyTorch (v2.1.2) libraries.

### E. COMPARATIVE METHODS

We compare our proposed method with a set of representative AES models including recent SOTA neural approaches.

**HISK** (ACL'18: short) [51] combines string kernels and word embeddings to capture essay features and uses support vector regression for scoring.

**STL-LSTM** (CoNLL'17) [7] builds a hierarchical sentence-document model that combines CNN and LSTM with an attention mechanism.

**MTL-BiLSTM** (NAACL'22) [52] utilizes a multi-task learning framework with bidirectional LSTM to score essays across multiple traits.

**ArTS** (EACL'24: findings) [45] introduces an autoregressive approach to multi-trait scoring by utilizing a pre-trained T5 model [43]. ArTS sequentially generates trait scores to capture inter-dependencies.

**ArTS+RMTS** (arXiv'24) [46] introduces Rationale-augmented Multi-Trait Scoring to ArTS, where rubric-guided rationales for traits are generated by LLMs and appended to ArTS inputs to improve the interpretability of AES.

**SaMRL** (EMNLP'24) [44] extends the ArTS model by incorporating Scoring-aware Multi-reward Reinforcement Learning, leveraging bidirectional QWK and trait-wise MSE rewards to enhance multi-trait AES.

**LLM baselines** [15] score essays with `Llama-3.1-8B-Instruct` in prompt-agnostic zero-/few-shot settings without task-specific fine-tuning. *LLM-abs* performs rubric-guided absolute scoring, while *LLM-comp* aggregates pairwise comparative judgments into a relative score; the numeric suffix denotes the number of in-context labeled examples (0 for zero-shot).

## V. RESULTS AND DISCUSSION

### A. LoRA AND SCORE ALIGNMENT IN FULL-DATA LEARNING

We evaluate our Two-Stage fine-tuning approach with LoRA, along with Score Alignment, applied to the baseline model DualBERT in the full-data setting. Our method is compared with existing AES approaches to analyze its effectiveness. UST was not applied in this experiment since unlabeled data is not available in our full-data experimental setup.

TABLE 3 and TABLE 4 present comparisons with existing methods in terms of per-trait and per-prompt performance, respectively. The performances of the existing methods were sourced from the papers [44], [46]. In our setting, the QWK of DualBERT differs from the original implementation [16], where our implementation shows a slightly higher QWK. This difference arises because we selected the model based on the average QWK of all traits, whereas Cho et al. [16] selected it based on the 'overall' trait alone. TABLE 3 and TABLE 4 indicate that DualBERT serves as a strong baseline in the multi-trait setting. While DualBERT exhibited slightly lower performance compared to ArTS+RMTS and SaMRL, it achieved better performance stability with a lower standard deviation. Since our study targets *multi-trait* scoring, selecting the checkpoint by the average QWK over all traits is the criterion that directly matches the evaluation objective, rather than optimizing a single trait. To quantify how much this choice affects our conclusions, TABLE 5 reports both criteria. Adopting the original 'overall'-trait criterion lowers the average QWK of DualBERT by about 0.02 (per-trait .701 vs. .679; per-prompt .720 vs. .701), but this gap narrows as our techniques are added (only .007 and .009 for +LoRA+SA). Even under the 'overall'-trait criterion, +LoRA+SA reaches .707/.723 (per-trait/per-prompt), on par with the best supervised baselines ArTS+RMTS (.708/.726) and SaMRL (.705/.722); under the average-of-traits criterion that matches the multi-trait objective, Score Alignment attains the best overall results.

LoRA improved the average QWK score by 0.6 points in both per-trait and per-prompt performances. Score Alignment, when applied to DualBERT, outperformed previous SOTA models—ArTS+RMTS and SaMRL—while demonstrating better performance stability (i.e., lower SD). Finally, the integration of LoRA and Score Alignment yielded additional performance gains, with QWK scores of 0.714 and 0.732 for per-trait and per-prompt performance, respectively. A paired $t$-test suggests these gains are statistically significant: over DualBERT, Score Alignment improves the AVG QWK by 0.011 (per-trait; 95% CI $[0.003, 0.019]$) and 0.008 (per-prompt; 95% CI $[0.003, 0.013]$), and LoRA+SA by 0.013 (per-trait; 95% CI $[0.007, 0.020]$) and 0.012 (per-prompt; 95% CI $[0.007, 0.017]$), with $p < 0.01$ for LoRA+SA. Qualitatively, these QWK values can be read against human consistency: the best results (.714 per trait and .732 per prompt) fall in the range conventionally interpreted as substantial agreement [53] and approach the human inter-rater agreement of 0.754 reported for overall scoring [17], indicating that the model's scores agree with human raters almost as closely as two human raters agree with each other.

The strength of LoRA and Score Alignment is that they can be easily integrated with deep learning–based regression models. However, for Seq2Seq-based models such as ArTS and SaMRL, which output in tokenized text format, applying Score Alignment can be difficult as it is designed to adjust the distribution of *continuous* predicted scores. As shown in TABLE 3 and TABLE 4, Score Alignment consistently improved performance across all traits from DualBERT. This result indicates that Score Alignment can provide additional performance improvements even in scenarios with sufficient training data.

### B. INTEGRATED METHOD IN K-DATA LEARNING

We evaluate our integrated method in a low-resource setting, specifically $K$-data learning scenarios, which utilize $K$ rated essays for each of the training and development sets. An ablation study is conducted by applying the three key techniques—Two-Stage fine-tuning using LoRA, Score Alignment, and UST—to the base model, DualBERT. In this setting, UST can additionally exploit the unlabeled pool described in Section IV-C; since this pool excludes the test set and all variants share the same labeled data, the comparison remains a fair ablation of each technique's contribution. In 'LoRA+SA+UST', Score Alignment is applied twice—both before and after UST. The first application helps generate better pseudo-labels for UST, and the second further refines the predicted scores. This experiment is designed to measure the contribution of each technique to the overall performance.

TABLE 6 and TABLE 7 present per-trait and per-prompt performance, respectively, in the *32*-data setting. ArTS+RMTS, which achieved SOTA performance in the *full*-data setting, showed degraded performance in the 32-data setting. We speculate that this is due to error propagation in the T5-based model, where insufficient training data leads to poor prediction of earlier scores, which in turn negatively affects the prediction of subsequent scores in the sequential decoding process. Concretely, ArTS-style models cast multi-trait scoring as text generation: the T5 decoder emits the trait scores one by one as tokens, each conditioned on the previously generated scores. During training, the decoder is conditioned on the gold score tokens, but at inference it must condition on its own predictions; with only 32 labeled essays, early-position predictions are frequently wrong, and these errors cascade along the autoregressive chain—an exposure-bias effect that compounds over the decoding sequence. Consistent with this mechanism, our reimplementation exhibits not only a large average drop but also a much larger variance across folds (SD ±.100 vs. ±.019 for DualBERT in TABLE 6), as occasional early errors can corrupt entire score sequences. In contrast, DualBERT predicts each trait with an independent regression head, so an error in one trait does not propagate to the others.

**TABLE 3.** Average QWK scores across prompts per trait in the *full*-data setting; SD denotes the averaged standard deviation for five seeds. Bold and underlined text indicate the best and second-best performances, respectively. * and ** denote $p < 0.05$ and $p < 0.01$, respectively, in a paired $t$-test of the AVG against DualBERT over five seeds.

| Method | Ovrl | Cnt | PA | Lng | Nar | Org | Cnv | WC | SF | Sty. | Voi. | AVG↑ (SD↓) |
|---|---|---|---|---|---|---|---|---|---|---|---|---|
| HISK [51] | .718 | .679 | .697 | .605 | .659 | .610 | .527 | .579 | .553 | .609 | .489 | .611 (-) |
| STL-LSTM [7] | .750 | .707 | .731 | .640 | .699 | .649 | .605 | .621 | .612 | .659 | .544 | .656 (-) |
| MTL-BiLSTM [52] | .764 | .685 | .701 | .604 | .668 | .615 | .560 | .615 | .598 | .632 | .582 | .638 (-) |
| ArTS [45] | .754 | .730 | .751 | .698 | .725 | .672 | .668 | .679 | .678 | .721 | .570 | .695 (±.018) |
| ArTS+RMTS [46] | .755 | .737 | **.752** | **.713** | **.744** | .682 | .690 | **.705** | .694 | .702 | .612 | .708 (±.043) |
| SaMRL [44] | .754 | .735 | .751 | .703 | .728 | .682 | .685 | .688 | .691 | .710 | **.627** | .705 (±.013) |
| DualBERT | .763 | .735 | .738 | .686 | .721 | .687 | .692 | .683 | .690 | .730 | .578 | .701 (±.012) |
| +LoRA | .779 | .742 | .750 | .682 | .722 | .700 | .688 | .685 | .691 | .719 | .620 | .707 (±.007) |
| +SA | .772 | .739 | .745 | .687 | .723 | .706 | .700 | .689 | .708 | **.743** | .613 | .712* (±.007) |
| +LoRA+SA | **.781** | **.748** | .751 | .688 | .726 | **.709** | **.704** | .687 | **.712** | .733 | .620 | **.714**** (±.011) |

**TABLE 4.** Average QWK scores across traits per prompt in the *full*-data setting; P1-8 denotes Prompt 1-8 and SD denotes the averaged standard deviation for five seeds. * and ** denote $p < 0.05$ and $p < 0.01$, respectively, in a paired $t$-test of the AVG against DualBERT over five seeds.

| Method | P1 | P2 | P3 | P4 | P5 | P6 | P7 | P8 | AVG↑ (SD↓) |
|---|---|---|---|---|---|---|---|---|---|
| HISK [51] | .674 | .586 | .651 | .681 | .693 | .709 | .641 | .516 | .644 (-) |
| STL-LSTM [7] | .690 | .622 | .663 | .729 | .719 | .753 | .704 | .592 | .684 (-) |
| MTL-BiLSTM [52] | .670 | .611 | .647 | .708 | .704 | .712 | .684 | .581 | .665 (-) |
| ArTS [45] | .708 | .706 | .704 | .767 | .723 | **.776** | .749 | .603 | .717 (±.025) |
| ArTS+RMTS [46] | .716 | .704 | **.723** | **.772** | **.737** | .769 | .736 | .651 | .726 (±.042) |
| SaMRL [44] | .702 | .711 | .708 | .766 | .722 | .773 | .743 | .649 | .722 (±.012) |
| DualBERT | .714 | .695 | .697 | .758 | .730 | .763 | .757 | .647 | .720 (±.009) |
| +LoRA | .728 | .709 | .705 | .760 | .731 | .766 | .755 | .656 | .726* (±.008) |
| +SA | .725 | .706 | .703 | .761 | .734 | .762 | **.768** | .666 | .728* (±.006) |
| +LoRA+SA | **.734** | **.713** | .710 | .764 | **.737** | .763 | .765 | **.674** | **.732**** (±.009) |

**TABLE 5.** Effect of the model-selection criterion in the *full*-data setting (average QWK over five seeds). "Ovrl-sel" selects the checkpoint by the 'overall' trait as in [16]; "Avg-sel" selects by the average QWK over all traits (ours).

| | Per-trait AVG | | Per-prompt AVG | |
|---|---|---|---|---|
| Method | Ovrl-sel | Avg-sel | Ovrl-sel | Avg-sel |
| DualBERT | .679 | .701 | .701 | .720 |
| +LoRA | .692 | .707 | .713 | .726 |
| +SA | .699 | .712 | .716 | .728 |
| +LoRA+SA | .707 | .714 | .723 | .732 |

In contrast, DualBERT maintains relatively high performance even in the low-resource setting. Furthermore, most of the performance values showed consistent improvements as each technique was applied. When scaled by 100, total improvement averaged 4.6 for per-trait QWK and 5.4 for per-prompt QWK. The final QWKs reached 0.639 for per-trait and 0.653 for per-prompt, corresponding to 91.2% and 90.7% of the full-data performance of DualBERT, i.e., 0.701 and 0.720, respectively. These results demonstrate the effectiveness of our integrated method in low-resource settings. In qualitative terms, DualBERT alone yields moderate agreement with human raters in this setting (QWK $\approx 0.6$), whereas the integrated method restores *substantial* agreement (QWK $> 0.61$ [53]) trained on only 32 rated essays—a level of scoring quality that remains practically meaningful under severe label scarcity. Paired $t$-tests suggest that these gains are statistically significant: LoRA+SA+UST improves the AVG QWK over DualBERT by 0.046 (per-trait; 95% CI [0.029, 0.063]) and 0.054 (per-prompt; 95% CI [0.038, 0.070]), both with $p < 0.01$, and Score Alignment alone is likewise significant ($p < 0.01$). In contrast, LoRA alone is not significant, and UST alone is significant only in the per-prompt setting. We attribute the insignificance of LoRA to the small development set: since the second stage of our Two-Stage fine-tuning selects the LoRA configuration based on development-set QWK, the few development samples available under the low-resource setting provide a noisy selection signal, which limits the benefit of LoRA. As a reference point, TABLE 6 also lists LLM baselines. On ASAP++, our LoRA+SA+UST (0.639) surpasses the strongest few-shot LLM baseline (LLM-abs-20, 0.603) despite using a far smaller model, and zero-shot LLMs remain substantially lower (0.26–0.48), consistent with the limited performance of LLM-based AES noted in the Introduction.

**TABLE 6.** Average QWK scores across prompts per trait in the 32-data setting. †We reimplemented ArTS+RMTS under the same experimental settings [46]. Zero-/few-shot LLM baselines (`Llama-3.1-8B-Instruct`) from [15] are shown for reference, with the suffix denoting the number of labeled examples. * and ** denote $p < 0.05$ and $p < 0.01$, respectively, in a paired $t$-test of the AVG against DualBERT over five folds.

| Method | Ovrl | Cnt | PA | Lng | Nar | Org | Cnv | WC | SF | Sty. | Voi. | AVG↑ (SD↓) |
|---|---|---|---|---|---|---|---|---|---|---|---|---|
| LLM-abs-0 | .252 | .414 | .250 | .454 | .257 | .141 | .133 | .353 | .384 | .080 | .137 | .259 (±.005) |
| LLM-comp-0 | .491 | .546 | .601 | .595 | .626 | .452 | .419 | .459 | .502 | .395 | .241 | .484 (±.011) |
| LLM-abs-5 | .560 | .584 | .459 | .599 | .546 | .520 | .447 | .593 | .625 | .459 | .438 | .530 (±.013) |
| LLM-abs-20 | .661 | .643 | .561 | .704 | .521 | .561 | .512 | .651 | .690 | .560 | .566 | .603 (±.007) |
| ArTS+RMTS† | .494 | .485 | .503 | .466 | .503 | .389 | .400 | .447 | .441 | .165 | .334 | .421 (±.100) |
| DualBERT | .636 | .605 | .600 | .544 | .573 | .598 | .598 | .595 | .611 | .613 | .549 | .593 (±.019) |
| +LoRA | .654 | .615 | .618 | .543 | .577 | .597 | .603 | .600 | .600 | .618 | .522 | .595 (±.018) |
| +SA | .677 | .631 | .648 | .601 | .628 | .625 | .626 | .616 | .624 | .673 | .575 | .630** (±.013) |
| +UST | .665 | .620 | .611 | .566 | .603 | .609 | .599 | .585 | .604 | .629 | .550 | .604 (±.010) |
| +LoRA+SA+UST | .686 | .650 | .668 | .626 | .651 | .634 | .641 | .625 | .629 | .686 | .528 | .639** (±.009) |
| **Total Imp. (×100)** | +5.0 | +4.5 | +6.8 | +8.2 | +7.8 | +3.6 | +4.3 | +3.0 | +1.8 | +7.3 | -2.1 | +4.6 |

**TABLE 7.** Average QWK scores across traits per prompt in the 32-data setting. †We reimplemented ArTS+RMTS under the same experimental settings [46]. * and ** denote $p < 0.05$ and $p < 0.01$, respectively, in a paired $t$-test of the AVG against DualBERT over five folds.

| Method | P1 | P2 | P3 | P4 | P5 | P6 | P7 | P8 | AVG↑ (SD↓) |
|---|---|---|---|---|---|---|---|---|---|
| ArTS+RMTS† | .479 | .416 | .494 | .562 | .543 | .475 | .374 | .349 | .462 (±.076) |
| DualBERT | .614 | .611 | .528 | .629 | .599 | .578 | .653 | .582 | .599 (±.018) |
| +LoRA | .621 | .617 | .544 | .651 | .605 | .577 | .661 | .576 | .607 (±.017) |
| +SA | .623 | .640 | .602 | .665 | .648 | .645 | .684 | .592 | .638** (±.014) |
| +UST | .625 | .613 | .557 | .653 | .619 | .608 | .665 | .576 | .615* (±.008) |
| +LoRA+SA+UST | .635 | .661 | .631 | .685 | .667 | .670 | .700 | .571 | .653** (±.011) |
| **Total Imp. (×100)** | +2.1 | +5.0 | +1.3 | +5.6 | +6.8 | +9.2 | +4.7 | -1.1 | +5.4 |

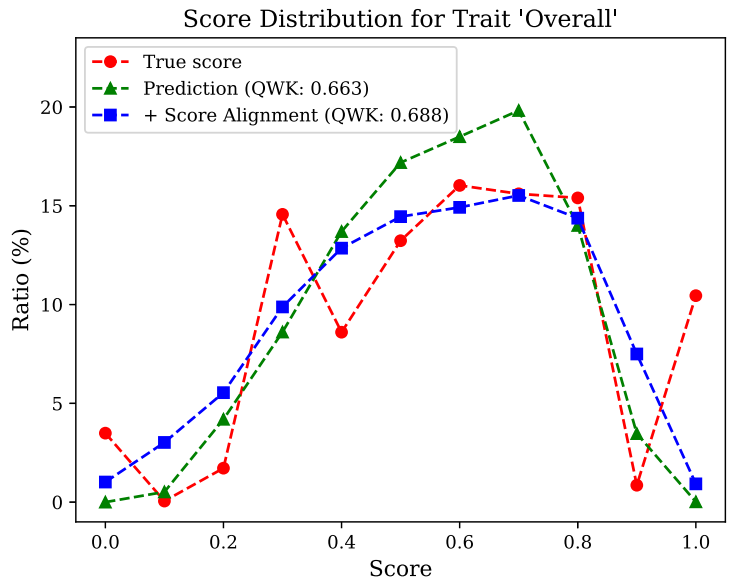

**FIGURE 2.** Effect of Score Alignment for trait 'overall'.

Among the applied techniques, Score Alignment yielded the most significant performance gains. Fig. 2 illustrates the shift in the predicted score distribution for the 'overall' trait before and after applying Score Alignment. Score Alignment improved the QWK score from 0.663 to 0.688. Fig. 2 demonstrates that Score Alignment adjusts the predicted score distribution to better align with the true score distribution. This result highlights that Score Alignment can mitigate deviations between the train and test domains.

### C. GENERALIZABILITY TO TOEFL11 AND ELLIPSE

We evaluate the proposed techniques on TOEFL11 and ELLIPSE. TABLE 8 reports the results, together with zero-/few-shot LLM baselines from [15] for reference. The trends are consistent with those on ASAP++. In the 32-data setting, Score Alignment and UST improve over DualBERT on both datasets, and the +SA+UST combination yields the largest gains: the ELLIPSE trait average rises from 0.484 to 0.530 and the TOEFL11 overall score from 0.514 to 0.586. In the full-data setting, Score Alignment improves the ELLIPSE trait average from 0.639 to 0.661, whereas the TOEFL11 overall score is already high ($\sim 0.725$) and changes little. LoRA alone brings only marginal change in these settings. As marked in TABLE 8, per-fold paired $t$-tests (baseline DualBERT, $n$=10) suggest that +SA, +UST, and +SA+UST significantly improve over DualBERT ($p < 0.05$) in the 32-data setting on both datasets, and that Score Alignment significantly improves the ELLIPSE trait average in the full-data setting ($p < 0.01$). Notably, unlike on ASAP++, LoRA slightly degrades performance on TOEFL11 and ELLIPSE in both the 32-data and full-data settings. We attribute this to their training configuration. Both datasets are trained in the multi-prompt (MP) setting, so the Two-Stage fine-tuning cannot exploit the per-trait, per-prompt loss-weight sweeping that drives its gains in the heterogeneous, single-prompt (SP) ASAP++ setup. Moreover, TOEFL11 has only a single overall trait and ELLIPSE has a more homogeneous trait set, leaving little trait-specific structure for the LoRA stage to adapt to; the additional LoRA parameters then tend to overfit rather than improve generalization.

On these two datasets, several zero-/few-shot LLM baselines are more competitive than on ASAP++: on TOEFL11, LLM-comp-0 (0.574) and LLM-abs-5 (0.579) approach our best low-resource model (+SA+UST, 0.586) while using fewer or no labeled examples. This apparent parity, however, comes with practical costs and caveats. First, the LLM baselines are computationally expensive at inference: LLM-comp relies on repeated pairwise comparisons and requires roughly one hour to score 100 essays, and LLM-abs-20 consumes roughly 10,000 tokens per sample on an 8B-scale model [15]. In contrast, DualBERT scores each essay (about 500 tokens) in a single forward pass and is far more compact: its parameter count is $(120.7 + 13.8\,n_{\text{trait}})$M, where $n_{\text{trait}}$ is the number of traits (about 134M and 217M for TOEFL11 and ELLIPSE, respectively), more than an order of magnitude smaller than an 8B-parameter LLM. Second, LLM accuracy does not scale reliably with additional labels: the best few-shot budget differs by dataset—20-shot on ASAP++ but 5-shot on TOEFL11 and ELLIPSE [15]—so supplying more labeled examples does not guarantee improvement. Therefore, even when a small amount of labeled data is available, LLM-based scoring is not unconditionally preferable, and our approach attains competitive or better QWK at a substantially lower inference cost, improving consistently both as each technique is added and as more labeled data become available (Fig. 3, on ASAP++).

**TABLE 8.** Average QWK scores on TOEFL11 and ELLIPSE (average over ten folds; SD in parentheses). Zero-/few-shot LLM baselines (`Llama-3.1-8B-Instruct`) are reported for reference, with the suffix denoting the number of labeled examples. The baseline performances are taken from [15]. * and ** denote $p < 0.05$ and $p < 0.01$, respectively, in a paired $t$-test against DualBERT over ten folds, computed on the TOEFL11 overall score and the ELLIPSE trait average.

| Dataset | TOEFL11 | ELLIPSE | | | | | | | |
|---|---|---|---|---|---|---|---|---|---|
| **Method** | **Ovrl** (SD↓) | **Ovrl** | **Coh** | **Syn** | **Voc** | **Phr** | **Grm** | **Cnv** | **Avg.↑** (SD↓) |
| *Zero-/Few-shot LLM Baselines*: `Llama-3.1-8B-Instruct` | | | | | | | | | |
| LLM-abs-0 | .395 (-) | .141 | .244 | .163 | .128 | .079 | .137 | .161 | .150 (-) |
| LLM-comp-0 | .574 (-) | **.526** | **.505** | **.493** | **.491** | **.484** | **.469** | **.504** | **.496** (-) |
| LLM-abs-5 | **.579** (-) | .477 | .418 | .436 | .434 | .424 | .418 | .418 | .432 (-) |
| LLM-abs-20 | .499 (-) | .403 | .321 | .373 | .333 | .414 | .388 | .345 | .368 (-) |
| *32-data Learning with DualBERT+{LoRA, SA, UST}* | | | | | | | | | |
| DualBERT | .514 (±.084) | .458 | .448 | .512 | .487 | .498 | .519 | .467 | .484 (±.055) |
| +LoRA | .511 (±.064) | .459 | .439 | .512 | .478 | .498 | .510 | .472 | .481 (±.050) |
| +SA | .558* (±.043) | .488 | .491 | .543 | .522 | .543 | .543 | .518 | .521* (±.039) |
| +UST | .574* (±.061) | .482 | .494 | .549 | **.530** | **.556** | .545 | .528 | .526** (±.043) |
| +SA+UST | **.586*** (±.043) | .491 | **.502** | **.552** | **.530** | .553 | **.546** | **.537** | **.530**** (±.046) |
| +LoRA+SA+UST | .574 (±.052) | **.496** | .499 | .548 | **.530** | .547 | .545 | .536 | .529* (±.034) |
| *Full-data Learning with DualBERT+{LoRA, SA, UST}* | | | | | | | | | |
| DualBERT | .725 (±.017) | .713 | .583 | .626 | .625 | .644 | .648 | .634 | .639 (±.009) |
| +LoRA | .724 (±.009) | .710 | .586 | .627 | .623 | .638 | .650 | .641 | .639 (±.007) |
| +SA | **.726** (±.016) | .723 | .618 | .651 | .640 | **.657** | .678 | .663 | .661** (±.006) |
| +LoRA+SA | .724 (±.009) | **.726** | **.619** | **.655** | **.645** | .656 | **.679** | **.666** | **.664**** (±.005) |

**TABLE 9.** Average QWK scores across traits per prompt in the 32-data setting on ASAP++, including results from the 5-runs and ensemble methods.

| Method | P1 | P2 | P3 | P4 | P5 | P6 | P7 | P8 | AVG↑ |
|---|---|---|---|---|---|---|---|---|---|
| (A): DualBERT | .614 | .611 | .528 | .629 | .599 | .578 | .653 | .582 | .599 |
| (B): (A)+5-runs | .612 | .625 | .529 | .639 | .601 | .596 | .643 | .563 | .601 |
| (C): (B)+LoRA | .619 | .627 | .537 | .644 | .613 | .618 | .648 | .570 | .610 |
| (D): (C)+Ens | .623 | .622 | .551 | .665 | .635 | .618 | .658 | .604 | .622 |
| (E): (D)+SA | .659 | .670 | .619 | .692 | .685 | .664 | .694 | .611 | .662 |
| (F): (E)+UST | .665 | .680 | .638 | .706 | .692 | .689 | .712 | .616 | .675 |
| **Total Imp.** | +5.1 | +6.9 | +11.0 | +7.7 | +9.3 | +11.1 | +5.9 | +3.4 | +7.6 |

### D. UPPER-BOUND PERFORMANCE ESTIMATION

The experiments in Sections V-D–V-H are conducted on ASAP++. To explore the upper-bound performance of our proposed methods, we apply two simple yet effective strategies: **5-runs**, where we select the best model out of five runs based on development set performance; and **Ensemble**, where four models are trained on different bootstrapped splits of the labeled data and combined via bagging. This experiment is designed to provide insight into the potential of our methods in practical settings. TABLE 9 and TABLE 10 present the ablation results in the 32-data and full-data settings, respectively.

It is well known that performance gains in deep learning tend to saturate as baseline performance increases; thus, achieving further improvements at high-performance levels is more challenging. However, this experiment demonstrates that the proposed methods can be applied orthogonally to simple performance-improving methods such as 5-runs and ensembling. In the 32-data setting, the total QWK score improved by 7.6 points, and even in the full-data setting, where performance is already strong, it improved by 2.4 points.

**TABLE 10.** Average QWK scores across traits per prompt in the *full*-data setting on ASAP++, including results from the 5-runs and ensemble methods.

| Method | P1 | P2 | P3 | P4 | P5 | P6 | P7 | P8 | AVG↑ |
|---|---|---|---|---|---|---|---|---|---|
| (A): DualBERT | .714 | .695 | .697 | .758 | .730 | .763 | .757 | .647 | .720 |
| (B): (A)+5-runs | .725 | .708 | .704 | .771 | .728 | .773 | .753 | .655 | .727 |
| (C): (B)+LoRA | .728 | .706 | .695 | .769 | .728 | .772 | .756 | .654 | .726 |
| (D): (C)+Ens | .735 | .719 | .712 | .778 | .739 | .777 | .765 | .671 | .737 |
| (E): (D)+SA | .746 | .727 | .718 | .777 | .742 | .776 | .773 | .696 | .744 |
| **Total Imp.** | +3.2 | +3.2 | +2.1 | +1.9 | +1.2 | +1.3 | +1.6 | +4.9 | +2.4 |

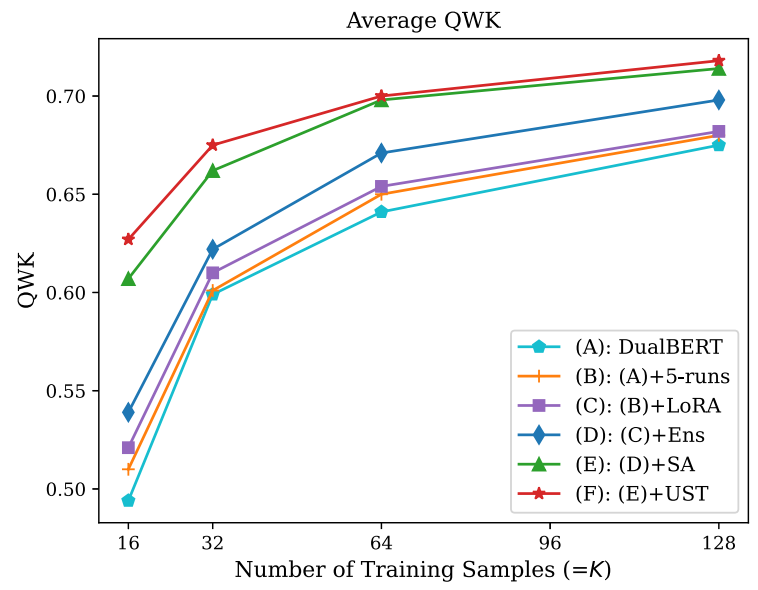

**FIGURE 3.** Effect of the number of training samples on ASAP++.

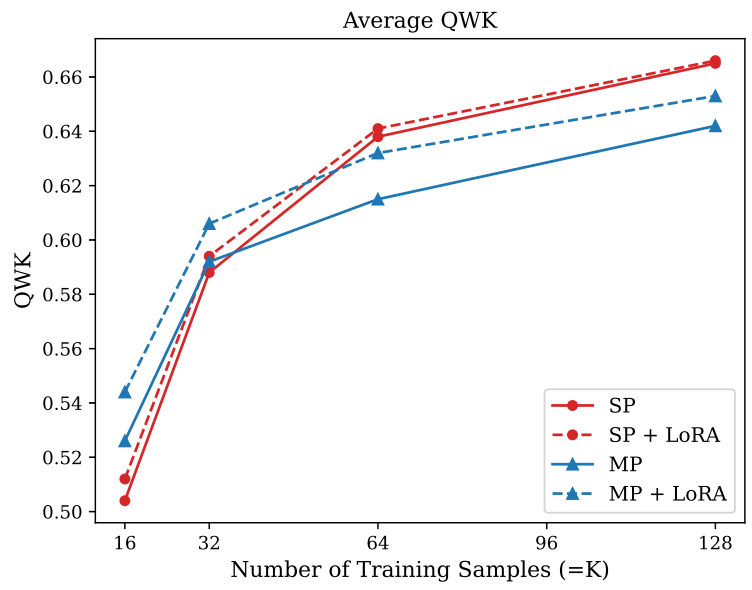

**FIGURE 4.** Effect of different prompt essays on ASAP++.

### *E. EFFECT OF THE NUMBER OF TRAINING SAMPLES*

We investigate how our methods' performance scales with varying training data size. Fig. 3 displays the performance curve according to the number of training samples (i.e., $K$). Performance refers to the average QWK across traits in the per-prompt performance. All techniques consistently improved performance across all sample sizes. Among them, Score Alignment shows the most substantial improvement.

When the number of training samples is small, the final integrated method (as denoted '(F): (E)+UST') significantly boosts the performance of DualBERT. However, this improvement becomes relatively small as the number of samples increases. Specifically, at $K=16$, the QWK improves from 0.494 to 0.627, an increase of approximately 0.133. In contrast, at $K=128$, the QWK improves from 0.675 to 0.718, showing a smaller increase of approximately 0.043. This result may be attributed to the robust generalization capability of deep learning.

### *F. EFFECT OF DIFFERENT PROMPT ESSAYS*

Essays from different prompts can either act as in-domain samples that enhance learning or as out-of-domain samples that lead to 'negative transfer' in multi-task learning. This section investigates the effect of LoRA in Two-Stage fine-tuning in the MP and SP settings (Section IV-C), as LoRA can function as a domain adapter across different prompts. In the MP setting, the model is trained on eight times more data ($\because$ eight prompts) than in the SP setting.

Fig. 4 displays the performance of LoRA in the MP and SP settings according to the number of training samples. We observe that MP outperforms SP when training data are very limited (e.g., 16 or 32 samples); however, above 64 samples, SP consistently yields better performance. At $K=128$, MP exhibits a QWK score approximately 0.04 points lower despite being trained on eight times more data. This suggests that in extremely low-resource settings, essays from different prompts act as in-domain samples. However, as the data size increases, they begin to function as out-of-domain samples, potentially causing negative transfer.

In addition, we can see that LoRA generally improves the performance in both settings. These results suggest that when training data for a *target* prompt are few (e.g., $K=16$) yet other prompts' data are sufficiently available, LoRA in Two-Stage fine-tuning can effectively serve as a domain adapter while leveraging the advantages of the MP setting.

### *G. EFFECT OF FILTERING NOISY LABELS IN UST*

In this experiment, we investigate how UST leverages unlabeled data to improve AES performance. UST uses the uncertainty measure to identify and exclude unreliable pseudo-labels of unrated essays in a self-training mechanism.

We further examine how the uncertainty estimate behaves and how sensitive it is to its configuration. Besides MC-dropout, which takes the standard deviation of predictions under repeated dropout, we consider an alternative estimator based on embedding noise: at each of the $T$ repetitions, Gaussian noise is added to the token embeddings of BERT, where the noise for each embedding dimension is scaled by the product of that dimension's standard deviation and a noise ratio $nr$. Fig. 5 reports both estimators at $T = 10$, comparing four subsets of the unlabeled set: the 256 most uncertain samples (Top-256), the 256 least uncertain samples (Bot-256), our balanced-low sampling (Bal-256), and the full set (All).

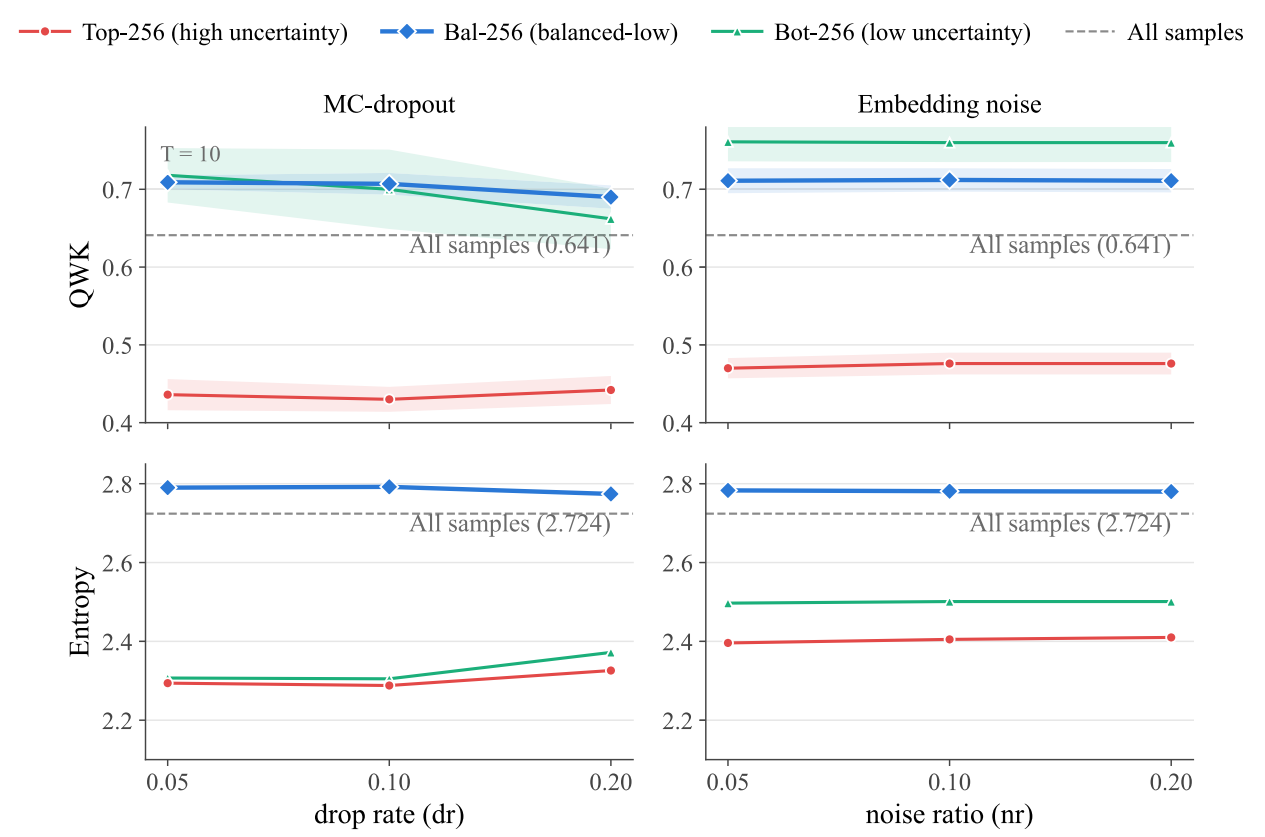

**FIGURE 5.** **Sensitivity of the uncertainty estimate at $T = 10$ on ASAP++. The top and bottom rows show the QWK and the score-distribution entropy of the selected subsets under MC-dropout (left, varying drop rate $dr$) and embedding noise (right, varying noise ratio $nr$). Top-256 and Bot-256 are the 256 most and least uncertain samples, Bal-256 is our balanced-low sampling (Section III-E), and the dashed line denotes the full unlabeled set.**

The uncertainty estimate clearly separates reliable from unreliable predictions: Top-256 yields a much lower QWK than Bot-256 and Bal-256, both of which exceed the full-set QWK. The two estimators show similar trends; at $dr = 0.1$ and $nr = 0.1$, the QWK gap between Bal-256 and Top-256 is about 0.28 for MC-dropout and 0.24 for embedding noise. Although Bot-256 attains a QWK comparable to Bal-256 on the selected subset, the entropy of the balanced subset's score distribution is closer to that of the full set (bottom row of Fig. 5); its pseudo-labels therefore better match the target score distribution and are more beneficial for self-training. Finally, both estimators are largely insensitive to their hyperparameters: varying $dr$ and $nr$ over $[0.05, 0.2]$ changes the results only marginally. Consequently, UST enhances AES performance by estimating uncertainty, filtering out noisy pseudo-labels, and training on a balanced set of low-uncertainty samples.

### *H. HYPERPARAMETER SENSITIVITY*

We examine the sensitivity of the proposed techniques to their key hyperparameters: the $p$ value of Score Alignment, the second-stage trait loss weights in Two-Stage fine-tuning, and the bin/sample sizes $(n_b, n_s)$ of UST. All experiments follow the *32*-data setting and report the per-trait AVG QWK as the mean and standard deviation over the five folds; TABLE 11 summarizes the results.

Score Alignment is robust to the choice of $p$: re-applying it with $p \in \{2.5, 5, 10, 20\}\%$ changes the AVG QWK by at most .004—less than the fold-level standard deviation—while its improvement over the unaligned predictions is preserved over the entire range. Even at $p$=2.5%, where the top and bottom anchor sets contain only two development essays in total, the drop from the default is .004.

**TABLE 11.** **Sensitivity of key hyperparameters in the *32*-data setting on ASAP++. Each value is the per-trait AVG QWK.**

| Configuration | AVG↑ (SD↓) |
|---|---|
| *p in Score Alignment* | |
| w/o SA | .593 (±.019) |
| $p = 2.5\%$ | .626 (±.011) |
| $p = 5\%$ (default) | .630 (±.013) |
| $p = 10\%$ | .630 (±.011) |
| $p = 20\%$ | .628 (±.010) |
| *Second-stage loss weights (with LoRA)* | |
| Alg. 1 weights (default) | .595 (±.018) |
| all trait weights $\sim U[0, 1]$ | .591 (±.014) |
| target $= 1$, others $\sim U[0, 1]$ | .590 (±.013) |
| *Bin/sample sizes $(n_b, n_s)$ in UST* | |
| w/o UST | .593 (±.019) |
| $(4, 64)$ | .604 (±.008) |
| $(8, 32)$ (default) | .604 (±.010) |
| $(16, 16)$ | .607 (±.009) |

For the loss weights in Two-Stage fine-tuning, we stress-test Alg. 1 by replacing the weights with random draws: either all trait weights are sampled from $U[0, 1]$, or the target trait's weight is fixed to 1 and the others are sampled. The two variants are nearly identical (difference $\leq$ .001) and stay within .005 of the default weights: since the second stage keeps only the configuration that performs best on the development set, the final performance does not hinge on the specific weight values. Consistent with TABLE 6, where the standalone gain of LoRA is small in this setting, all three configurations remain within one fold-level standard deviation of DualBERT.

UST is likewise insensitive to how the budget of $n_b \times n_s = 256$ pseudo-labels is partitioned (difference $\leq$ .003). Together with the analysis of the uncertainty estimator in Section V-G (Fig. 5), these results suggest that, within the examined ranges, the three techniques are relatively insensitive to their hyperparameter settings.

## VI. LIMITATIONS

Our methods, including Two-Stage fine-tuning, Score Alignment, and UST, have several limitations as follows. First, the Two-Stage fine-tuning strategy with LoRA requires more computational time for training than its base model. As shown in Alg. 1, the training complexity increases linearly with the number of traits. However, the inference time remains almost the same, and the increased training time is not significant for a small dataset. When scaling to rubrics with many more traits, the second stage requires $|\mathbb{T}_p|+1$ LoRA runs (the 'balance' configuration plus one per trait), but each run updates only the adapter layers on a shared frozen base, so the per-run cost is low and the runs are mutually independent—they can be executed in parallel on separate devices, keeping the wall-clock time nearly constant given sufficient GPUs. For very large trait sets, we recommend two concrete strategies: (i) sweeping only the 'balance' and 'overall' configurations plus target-specific configurations for the traits of primary interest, and (ii) grouping strongly correlated traits so that each group shares a single loss-weight configuration, which reduces the number of runs from $|\mathbb{T}_p|+1$ to the number of groups. Second, Score Alignment is a post-processing method that requires generating predictions for the entire test (or unlabeled) set before application. Third, UST relies on access to a large amount of unrated essays that follow the target prompt's distribution, which may not always be available in real-world applications. Nevertheless, our method can still be valuable in settings that require grading a large volume of student essays, and its balanced low-uncertainty sampling can partially mitigate the effect of a distribution shift that may arise in the test domain. Collecting a large number of unrated essays from diverse external sources could address the second and third limitations. In future work, we plan to develop a zero-data learning method to address these constraints.

## VII. CONCLUSION

Data scarcity poses significant challenges to building robust AES systems in real-world settings. In this study, we propose a novel AES approach by introducing three key techniques. Two-Stage fine-tuning with LoRA improves prediction performance, applying distinct weights for each trait during training. Score Alignment improves the consistency between predicted and true score distributions through a linear transformation based on predictions of the development set. UST reduces noisy pseudo-labels by filtering out uncertain samples during self-training.

Our experiments on the ASAP++ dataset demonstrate the effectiveness of the proposed approach. In the *32*-data setting, the integrated method improved QWK by 4.6 points from the base model, achieving 91.2% of its full-data performance. In the full-data setting, Score Alignment combined with DualBERT achieved new state-of-the-art performance, outperforming existing AES methods.

This study highlights the effectiveness of our AES approach in both few- and full-data settings. In the future, we plan to develop a zero-data learning method for real-world applications.

## ACKNOWLEDGMENT

This study was conducted while she was with ETRI.

## REFERENCES

[1] N. Almusharraf and H. Alotaibi, "An error-analysis study from an EFL writing context: Human and automated essay scoring approaches," *Technol., Knowl. Learn.*, vol. 28, no. 3, pp. 1015–1031, Sep. 2023, doi: 10.1007/s10758-022-09592-z.

[2] A. P. Cavalcanti, A. Barbosa, R. Carvalho, F. Freitas, Y.-S. Tsai, D. Gašević, and R. F. Mello, "Automatic feedback in online learning environments: A systematic literature review," *Comput. Educ., Artif. Intell.*, vol. 2, Jul. 2021, Art. no. 100027.

[3] D. Ifenthaler, "Automated essay scoring systems," in *Handbook of Open, Distance and Digital Education*. Cham, Switzerland: Springer, 2023, pp. 1057–1071.

[4] E. D. Reilly, R. E. Stafford, K. M. Williams, and S. B. Corliss, "Evaluating the validity and applicability of automated essay scoring in two massive open online courses," *Int. Rev. Res. Open Distrib. Learn.*, vol. 15, no. 5, pp. 83–98, Oct. 2014.

[5] D. Ramesh and S. K. Sanampudi, "An automated essay scoring systems: A systematic literature review," *Artif. Intell. Rev.*, vol. 55, no. 3, pp. 2495–2527, Mar. 2022.

[6] W. Xu, R. Mahmud, and W. L. Hoo, "A systematic literature review: Are automated essay scoring systems competent in real-life education scenarios?" *IEEE Access*, vol. 12, pp. 77639–77657, 2024.

[7] F. Dong, Y. Zhang, and J. Yang, "Attention-based recurrent convolutional neural network for automatic essay scoring," in *Proc. 21st Conf. Comput. Natural Lang. Learn. (CoNLL)*, Aug. 2017, pp. 153–162. [Online]. Available: https://aclanthology.org/K17-1017

[8] R. Yang, J. Cao, Z. Wen, Y. Wu, and X. He, "Enhancing automated essay scoring performance via fine-tuning pre-trained language models with combination of regression and ranking," in *Proc. Findings Assoc. Comput. Linguistics: EMNLP*, Nov. 2020, pp. 1560–1569. [Online]. Available: https://aclanthology.org/2020.findings-emnlp.141

[9] Y. Wang, C. Wang, R. Li, and H. Lin, "On the use of BERT for automated essay scoring: Joint learning of multi-scale essay representation," in *Proc. Conf. North Amer. Chapter Assoc. Comput. Linguistics, Hum. Lang. Technol.*, Jul. 2022, pp. 3416–3425. [Online]. Available: https://aclanthology.org/2022.naacl-main.249

[10] T. Brown et al., "Language models are few-shot learners," in *Proc. Adv. Neural Inf. Process. Syst.*, vol. 33, 2020, pp. 1877–1901.

[11] W. X. Zhao, K. Zhou, J. Li, T. Tang, X. Wang, Y. Hou, Y. Min, B. Zhang, J. Zhang, Z. Dong, Y. Du, C. Yang, Y. Chen, Z. Chen, J. Jiang, R. Ren, Y. Li, X. Tang, Z. Liu, P. Liu, J.-Y. Nie, and J.-R. Wen, "A survey of large language models," 2023, *arXiv:2303.18223*.

[12] S. Lee, Y. Cai, D. Meng, Z. Wang, and Y. Wu, "Unleashing large language models' proficiency in zero-shot essay scoring," in *Proc. Findings Assoc. Comput. Linguistics, (EMNLP)*, Nov. 2024, pp. 181–198. [Online]. Available: https://aclanthology.org/2024.findings-emnlp.10

[13] M. Stahl, L. Biermann, A. Nehring, and H. Wachsmuth, "Exploring LLM prompting strategies for joint essay scoring and feedback generation," in *Proc. 19th Workshop Innov. Use NLP Building Educ. Appl. (BEA)*, Jun. 2024, pp. 283–298. [Online]. Available: https://aclanthology.org/2024.bea-1.23

[14] L. Wang, X. Chen, C. Wang, L. Xu, R. Shadiev, and Y. Li, "ChatGPT's capabilities in providing feedback on undergraduate students' argumentation: A case study," *Thinking Skills Creativity*, vol. 51, Mar. 2024, Art. no. 101440.

[15] H. Choi, M.-C. Kang, J. Seong, and J.-X. Huang, "Exploring zero-shot essay scoring: From feature-based to LLM-based approaches," *Data Mining Knowl. Discovery*, vol. 40, no. 3, p. 35, May 2026.

[16] M. Cho, J.-X. Huang, and O.-W. Kwon, "Dual-scale BERT using multi-trait representations for holistic and trait-specific essay grading," *ETRI J.*, vol. 46, no. 1, pp. 82–95, Feb. 2024.

[17] K. Taghipour and H. T. Ng, "A neural approach to automated essay scoring," in *Proc. Conf. Empirical Methods Natural Lang. Process.*, Nov. 2016, pp. 1882–1891. [Online]. Available: https://aclanthology.org/D16-1193

[18] X. Li, M. Chen, and J.-Y. Nie, "SEDNN: Shared and enhanced deep neural network model for cross-prompt automated essay scoring," *Knowledge-Based Syst.*, vol. 210, Dec. 2020, Art. no. 106491. [Online]. Available: https://www.sciencedirect.com/science/article/pii/S0950705120306201

[19] R. Ridley, L. He, X.-Y. Dai, S. Huang, and J. Chen, "Automated cross-prompt scoring of essay traits," in *Proc. AAAI Conf. Artif. Intell.*, May 2021, vol. 35, no. 15, pp. 13745–13753. [Online]. Available: https://ojs.aaai.org/index.php/AAAI/article/view/17620

[20] H. Do, Y. Kim, and G. G. Lee, "Prompt- and trait relation-aware cross-prompt essay trait scoring," in *Proc. Findings Assoc. Comput. Linguistics: ACL*. Toronto, ON, Canada: Association for Computational Linguistics, Jul. 2023, pp. 1538–1551. [Online]. Available: https://aclanthology.org/2023.findings-acl.98

[21] S. Li and V. Ng, "Conundrums in cross-prompt automated essay scoring: Making sense of the state of the art," in *Proc. 62nd Annu. Meeting Assoc. Comput. Linguistics*, Aug. 2024, pp. 7661–7681. [Online]. Available: https://aclanthology.org/2024.acl-long.414

[22] T. Shibata and M. Uto, "Cross-prompt automated essay scoring via reinforcement learning-based data valuation," *IEEE Access*, vol. 13, pp. 184792–184808, 2025.

[23] Q. Tao, J. Zhong, and R. Li, "AESPrompt: Self-supervised constraints for automated essay scoring with prompt tuning," in *Proc. Int. Conf. Softw. Eng. Knowl. Eng.*, vol. 2022, Jul. 2022, pp. 335–340.

[24] J. He and X. Li, "Zero-shot cross-lingual automated essay scoring," in *Proc. Joint Int. Conf. Comput. Linguistics, Lang. Resour. Eval.*, May 2024, pp. 17819–17832. [Online]. Available: https://aclanthology.org/2024.lrec-main.1550

[25] E. J. Hu, Y. Shen, P. Wallis, Z. Allen-Zhu, Y. Li, S. Wang, L. Wang, and W. Chen, "LoRA: Low-rank adaptation of large language models," 2021, *arXiv:2106.09685*.

[26] S. Mukherjee and A. Awadallah, "Uncertainty-aware self-training for few-shot text classification," in *Proc. Adv. Neural Inf. Process. Syst.*, vol. 33, 2020, pp. 21199–21212.

[27] S. Mathias and P. Bhattacharyya, "ASAP++: Enriching the ASAP automated essay grading dataset with essay attribute scores," in *Proc. 11th Int. Conf. Lang. Resour. Eval. (LREC)*, May 2018, pp. 1169–1173. [Online]. Available: https://aclanthology.org/L18-1187

[28] D. Blanchard, J. Tetreault, D. Higgins, A. Cahill, and M. Chodorow, "TOEFL11: A corpus of non-native English," *ETS Res. Rep. Ser.*, vol. 2013, no. 2, pp. 1–15, Dec. 2013.

[29] S. Crossley, Y. Tian, P. Baffour, A. Franklin, Y. Kim, W. Morris, M. Benner, A. Picou, and U. Boser, "The English language learner insight, proficiency and skills evaluation (ELLIPSE) corpus," *Int. J. Learner Corpus Res.*, vol. 9, no. 2, pp. 248–269, Dec. 2023.

[30] P. Phandi, K. M. A. Chai, and H. T. Ng, "Flexible domain adaptation for automated essay scoring using correlated linear regression," in *Proc. Conf. Empirical Methods Natural Lang. Process.* Lisbon, Portugal: Association for Computational Linguistics, Sep. 2015, pp. 431–439. [Online]. Available: https://aclanthology.org/D15-1049

[31] Y. Tay, M. Phan, L. A. Tuan, and S. C. Hui, "Skipflow: Incorporating neural coherence features for end-to-end automatic text scoring," in *Proc. AAAI Conf. Artif. Intell.*, vol. 32, 2018, pp. 5948–5955.

[32] Y. Cao, H. Jin, X. Wan, and Z. Yu, "Domain-adaptive neural automated essay scoring," in *Proc. 43rd Int. ACM SIGIR Conf. Res. Develop. Inf. Retr.* New York, NY, USA: Association for Computing Machinery, Jul. 2020, pp. 1011–1020, doi: 10.1145/3397271.3401037.

[33] W. Song, Z. Song, L. Liu, and R. Fu, "Hierarchical multi-task learning for organization evaluation of argumentative student essays," in *Proc. 29th Int. Joint Conf. Artif. Intell.*, Jul. 2020, pp. 3875–3881.

[34] D. Liao, J. Xu, G. Li, and Y. Wang, "Hierarchical coherence modeling for document quality assessment," in *Proc. AAAI Conf. Artif. Intell.*, May 2021, vol. 35, no. 15, pp. 13353–13361. [Online]. Available: https://ojs.aaai.org/index.php/AAAI/article/view/17576

[35] M. Uto, "A review of deep-neural automated essay scoring models," *Behaviormetrika*, vol. 48, no. 2, pp. 459–484, Jul. 2021.

[36] J. Xie, K. Cai, L. Kong, J. Zhou, and W. Qu, "Automated essay scoring via pairwise contrastive regression," in *Proc. 29th Int. Conf. Comput. Linguistics*, Oct. 2022, pp. 2724–2733. [Online]. Available: https://aclanthology.org/2022.coling-1.240

[37] Z. Jiang, T. Gao, Y. Yin, M. Liu, H. Yu, Z. Cheng, and Q. Gu, "Improving domain generalization for prompt-aware essay scoring via disentangled representation learning," in *Proc. 61st Annu. Meeting Assoc. Comput. Linguistics*. Toronto, ON, Canada: Association for Computational Linguistics, Jul. 2023, pp. 12456–12470. [Online]. Available: https://aclanthology.org/2023.acl-long.696

[38] C. Jin, B. He, K. Hui, and L. Sun, "TDNN: A two-stage deep neural network for prompt-independent automated essay scoring," in *Proc. 56th Annu. Meeting Assoc. Comput. Linguistics*. Melbourne, VIC, Australia: Association for Computational Linguistics, Jul. 2018, pp. 1088–1097. [Online]. Available: https://aclanthology.org/P18-1100

[39] J. Devlin, M.-W. Chang, K. Lee, and K. Toutanova, "BERT: Pre-training of deep bidirectional transformers for language understanding," in *Proc. Conf. North Amer. Chapter Assoc. Comput. Linguistics, Hum. Lang. Technol.*, vol. 1, Jun. 2019, pp. 4171–4186. [Online]. Available: https://aclanthology.org/N19-1423

[40] Y. Liu, "RoBERTa: A robustly optimized BERT pretraining approach," 2019, *arXiv:1907.11692*.

[41] J. Xue, X. Tang, and L. Zheng, "A hierarchical BERT-based transfer learning approach for multi-dimensional essay scoring," *IEEE Access*, vol. 9, pp. 125403–125415, 2021.

[42] A. Sethi and K. Singh, "Natural language processing based automated essay scoring with parameter-efficient transformer approach," in *Proc. 6th Int. Conf. Comput. Methodologies Commun. (ICCMC)*, Mar. 2022, pp. 749–756.

[43] C. Raffel, N. Shazeer, A. Roberts, K. Lee, S. Narang, M. Matena, Y. Zhou, W. Li, and P. J. Liu, "Exploring the limits of transfer learning with a unified text-to-text transformer," *J. Mach. Learn. Res.*, vol. 21, no. 140, pp. 1–67, 2020.

[44] H. Do, S. Ryu, and G. Lee, "Autoregressive multi-trait essay scoring via reinforcement learning with scoring-aware multiple rewards," in *Proc. Conf. Empirical Methods Natural Lang. Process.*, Nov. 2024, pp. 16427–16438. [Online]. Available: https://aclanthology.org/2024.emnlp-main.917

[45] H. Do, Y. Kim, and G. Lee, "Autoregressive score generation for multi-trait essay scoring," in *Proc. Findings Assoc. Comput. Linguistics: EACL*, Mar. 2024, pp. 1659–1666. [Online]. Available: https://aclanthology.org/2024.findings-eacl.115/

[46] S. Chu, J. Kim, B. Wong, and M. Yi, "Rationale behind essay scores: Enhancing S-LLM's multi-trait essay scoring with rationale generated by LLMs," 2024, *arXiv:2410.14202*.

[47] A. Mizumoto and M. Eguchi, "Exploring the potential of using an AI language model for automated essay scoring," *Res. Methods Appl. Linguistics*, vol. 2, no. 2, Aug. 2023, Art. no. 100050. [Online]. Available: https://www.sciencedirect.com/science/article/pii/S2772766123000101

[48] Y. Song, Q. Zhu, H. Wang, and Q. Zheng, "Automated essay scoring and revising based on open-source large language models," *IEEE Trans. Learn. Technol.*, vol. 17, pp. 1880–1890, 2024.

[49] I. Loshchilov and F. Hutter, "Decoupled weight decay regularization," in *Proc. Int. Conf. Learn. Represent.*, 2019. [Online]. Available: https://openreview.net/forum?id=Bkg6RiCqY7

[50] N. Srivastava, G. Hinton, A. Krizhevsky, I. Sutskever, and R. Salakhutdinov, "Dropout: A simple way to prevent neural networks from overfitting," *J. Mach. Learn. Res.*, vol. 15, no. 56, pp. 1929–1958, 2014. [Online]. Available: http://jmlr.org/papers/v15/srivastava14a.html

[51] M. Cozma, A. Butnaru, and R. T. Ionescu, "Automated essay scoring with string kernels and word embeddings," in *Proc. 56th Annu. Meeting Assoc. Comput. Linguistics (Volume 2: Short Papers)*. Melbourne, Australia: Association for Computational Linguistics, Jul. 2018, pp. 503–509. [Online]. Available: https://aclanthology.org/P18-2080

[52] R. Kumar, S. Mathias, S. Saha, and P. Bhattacharyya, "Many hands make light work: Using essay traits to automatically score essays," in *Proc. Conf. North Amer. Chapter Assoc. Comput. Linguistics, Hum. Lang. Technol.*, Jul. 2022, pp. 1485–1495. [Online]. Available: https://aclanthology.org/2022.naacl-main.106

[53] J. R. Landis and G. G. Koch, "The measurement of observer agreement for categorical data," *Biometrics*, vol. 33, no. 1, pp. 159–174, Mar. 1977.

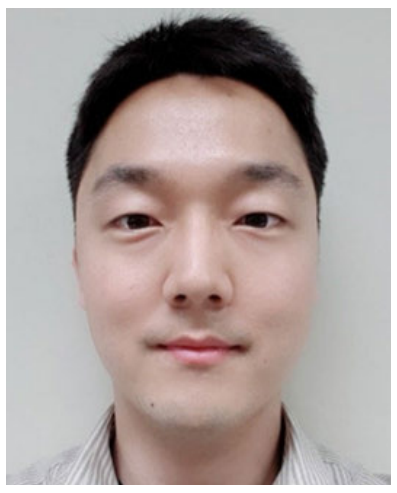

**HONGSEOK CHOI** received the B.S. degree in information and communication engineering from Inha University, Incheon, South Korea, in 2016, and the Ph.D. degree in electrical engineering and computer science from Gwangju Institute of Science and Technology (GIST), Gwangju, South Korea, in 2023.

He is currently a Senior Researcher with the Electronics and Telecommunications Research Institute (ETRI), Daejeon, South Korea. His research interests include natural language processing and multi-agent orchestration.

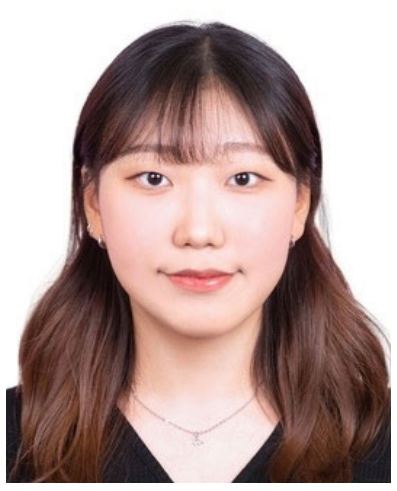

**SERYNN KIM** is currently pursuing the B.S. degrees in English linguistics and language technology (ELLT) and artificial intelligence convergence with Hankuk University of Foreign Studies, Seoul, South Korea. She was a Research Intern with the Electronics and Telecommunications Research Institute (ETRI), Daejeon, South Korea, in 2025. Her research interests include natural language processing and large language models.

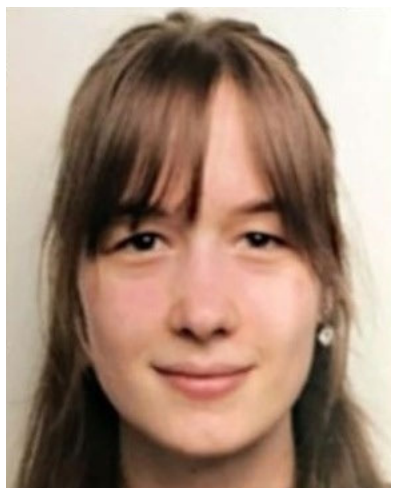

**WENCKE LIERMANN** received the B.S. degree in computational linguistics from the University of Potsdam, Potsdam, Germany, in 2022, and the M.S. degree in artificial intelligence from Chungnam National University, Daejeon, South Korea, in 2024. Since 2024, she has been a Post-Master Research Fellow with the Language Intelligence Research Section, Electronics and Telecommunications Research Institute (ETRI), Daejeon, South Korea.

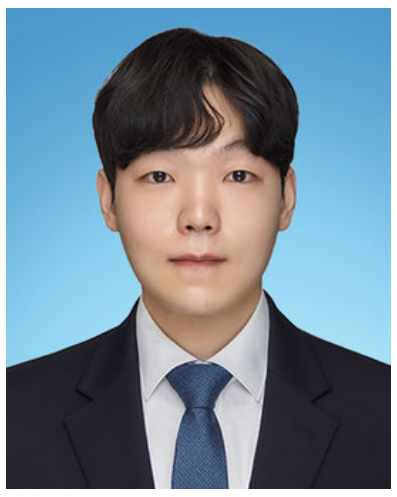

**JIN SEONG** received the B.S. degree in computer science from Kwangwoon University, Seoul, South Korea, in 2020, and the M.S. degree in computer science from Hanyang University, Seoul, South Korea, in 2022. He is currently a Researcher with the Electronics and Telecommunications Research Institute (ETRI), Daejeon, South Korea. His research interests include vision-language models (VLMs) and natural language processing.

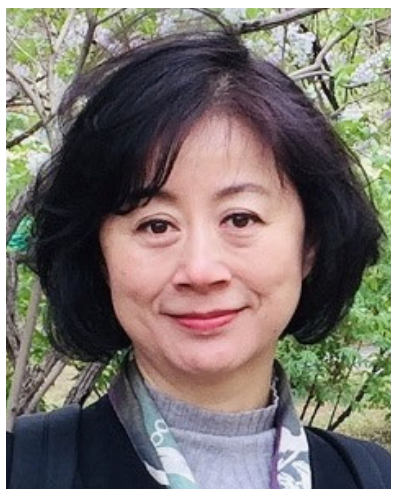

**JIN-XIA HUANG** received the B.S. degree in physics from Jilin University, China, in 1991, the M.S. degree in computer science from KAIST, South Korea, in 2001, and the Ph.D. degree in computer science from Jeonbuk National University, South Korea, in 2018.

From 1994 to 1997, she was an Engineer with Yanbian University of Science and Technology (YUST), China, and from 2001 to 2003, she was a Researcher with Microsoft Research Asia (MSRA), China. Since 2008, she has been with the Electronics and Telecommunications Research Institute (ETRI), South Korea, where she is currently a Principal Researcher with the Language Intelligence Research Section. Since March 2024, she has been an Associate Professor in AI Major with UST ETRI School. Her research has focused on natural language processing, including dialogue systems and intelligent tutoring agents. Her current research interests include intelligent multi-agent orchestration and AI-driven scientific discovery.